\documentclass[letterpaper, 10 pt, conference]{ieeeconf}  

\IEEEoverridecommandlockouts                              
\usepackage{multirow}
\usepackage{amsmath} 
\usepackage{amssymb}  
\usepackage{algorithm}
\usepackage{amsfonts}
\usepackage{algorithmic}
\usepackage{booktabs} 
\usepackage{siunitx} 
\usepackage[utf8]{inputenc}
\usepackage[T1]{fontenc}
\usepackage{lmodern}
\usepackage{graphicx}
\graphicspath{{Figure/}}                
\DeclareGraphicsExtensions{.pdf,.png,.jpg}
\usepackage[section]{placeins}
\usepackage{float}
\usepackage{subcaption}

\title{\LARGE \bf
A Mobile Magnetic Manipulation Platform for Gastrointestinal Navigation with Deep Reinforcement Learning Control
}

\author{%
  Zhifan Yan$^{1,*}$,
  Chang Liu$^{1,*}$,
  Yiyang Jiang$^{1,*}$,
  Wenxuan Zheng$^{1,*}$,
  Xinhao Chen$^{1}$,
  Axel Krieger$^{1}$%
  \thanks{* These authors contributed equally to this work.}%
  \thanks{$^{1}$ Laboratory for Computational Sensing and Robotics (LCSR), Johns Hopkins University, Baltimore, MD 21218, USA. Emails: \texttt{zyan41@jhu.edu}.}%
}

\begin{document}

\maketitle
\thispagestyle{empty}
\pagestyle{empty}

\begin{abstract}
Targeted drug delivery in the gastrointestinal (GI) tract using magnetic robots offers a promising alternative to systemic treatments. However, controlling these robots is a major challenge. Stationary magnetic systems have a limited workspace, while mobile systems (e.g., coils on a robotic arm) suffer from a "model-calibration bottleneck", requiring complex, pre-calibrated physical models that are time-consuming to create and computationally expensive. This paper presents a compact, low-cost mobile magnetic manipulation platform that overcomes this limitation using Deep Reinforcement Learning (DRL). Our system features a compact four-electromagnet array mounted on a UR5 collaborative robot. A Soft Actor–Critic (SAC)–based control strategy is trained through a sim-to-real pipeline, enabling effective policy deployment within 15 minutes and significantly reducing setup time. We validated the platform by controlling a 7-mm magnetic capsule along 2D trajectories. Our DRL-based controller achieved a root-mean-square error (RMSE) of 1.18~mm for a square path and 1.50~mm for a circular path. We also demonstrated successful tracking over a clinically relevant, 30 cm $\times$ 20 cm workspace. This work demonstrates a rapidly deployable, model-free control framework capable of precise magnetic manipulation in a large workspace,validated using a 2D GI phantom.
\end{abstract}

\section*{Introduction}
\begin{figure*}[t]
  \centering
  \includegraphics[width=0.9\textwidth]{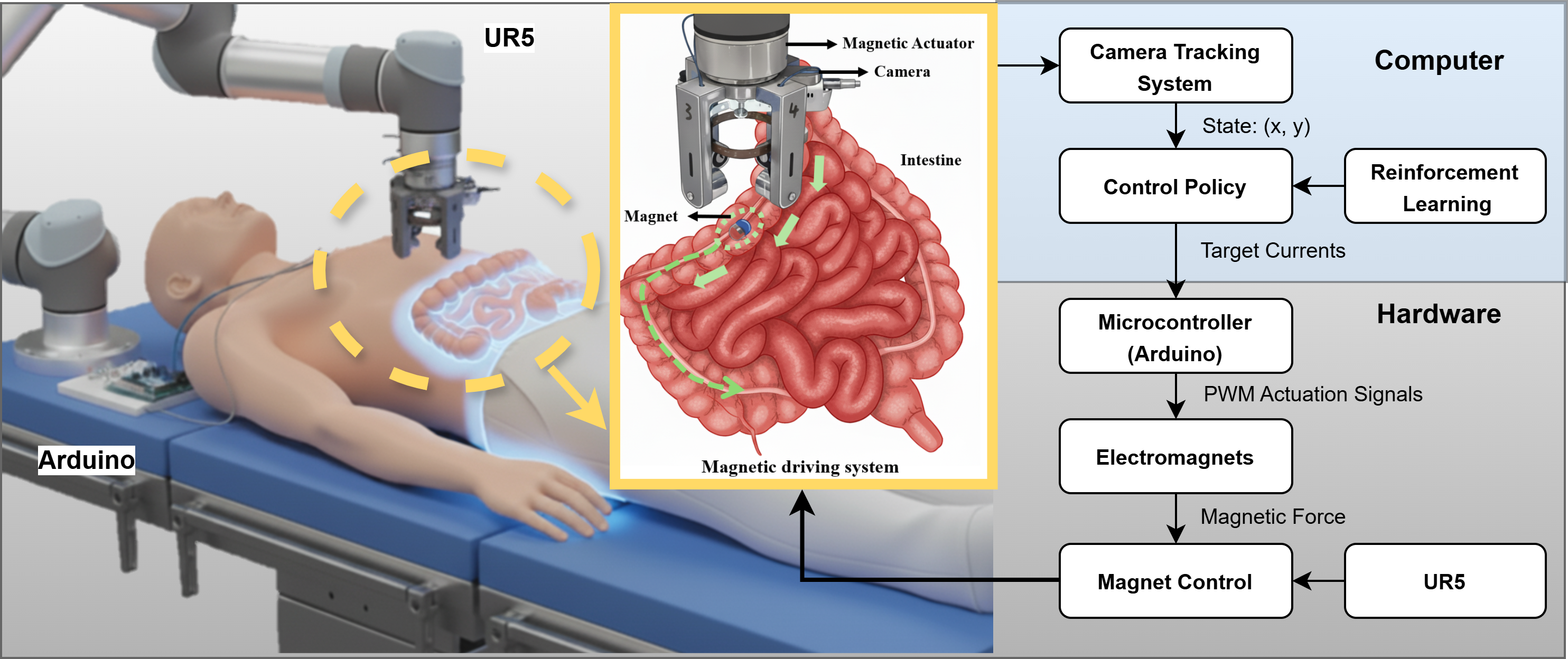}
  \caption{Overview of our mobile magnetic manipulation platform and control system.}
  \label{fig:overview}
\end{figure*}

Gastrointestinal (GI) diseases, like colorectal cancer (CRC), are a significant global health burden \cite{sung2021global}. Current treatments such as systemic chemotherapy suffer from severe dose-limiting toxicities (DLTs) \cite{pantoja2022designing}, while conventional oral drugs face formidable physiological barriers (e.g., pH variations, enzymatic degradation) that limit bioavailability \cite{alqahtani2021advances, hua2020advances}. Targeted drug delivery, which precisely transports therapeutics to a lesion, is a compelling solution to enhance efficacy and minimize off-target toxicity. However, reliably navigating the complex GI tract to achieve this remains a major challenge.

Untethered magnetic microrobots represent a promising paradigm for non-invasive navigation and intervention within the GI tract. These robots can be specifically engineered to minimize off-target drug release; for example, recent strategies include pH-sensitive shells for gastric protection and anchoring mechanisms for precise therapeutic seeding in the intestine \cite{chen2021triple}. However, achieving reliable in-vivo actuation of these robots remains a major challenge.

Magnetic actuation systems have evolved from large, stationary coil arrays to mobile platforms. Stationary systems, including advanced multi-coil designs, can offer high-fidelity control but are fundamentally constrained by a small effective workspace, often limited to the decimeter-scale \cite{erin2024strong}. This limited volume and high infrastructure cost restrict clinical applicability for navigating the large and tortuous GI tract. To address this workspace limitation, researchers have mounted magnetic sources on robotic arms, as seen in platforms like ARMM \cite{sikorski2019armm} and RoboMag \cite{du2020robomag,davy2023independent}. While this paradigm theoretically expands the workspace to the reach of the manipulator, it introduces a substantial new control challenge: the magnetic field dynamics become directly coupled with the robot's pose, creating a highly complex, time-varying, and non-linear control problem.

Existing approaches for controlling these mobile systems suffer from what we term a "model-calibration bottleneck": the requirement for time-consuming (hours to days) pre-calibration of precise magnetic field models via finite element analysis or extensive measurements to compute coil currents \cite{sikorski2019armm, du2020robomag, davy2023independent, chen2024mitigating}. These models are computationally expensive to evaluate in real-time, hindering deployment in dynamic environments. While reinforcement learning (RL) can bypass explicit modeling \cite{kober2013reinforcement, salehi2024intelligent}, its application in complex systems is often hindered by long training times and the sim-to-real transfer gap \cite{barnoy2022control}.

In this work, we address this bottleneck by introducing a portable, low-cost magnetic manipulation platform that replaces model-based control with a learning-based framework(see Fig.~\ref{fig:overview}). Our hardware comprises a compact four-electromagnet array mounted as an end-effector on a UR5 collaborative robot. By leveraging the arm’s mobility, the coils can be positioned near the target, producing strong, localized magnetic field gradients at low power while providing a large, reconfigurable workspace. This architecture decouples electromagnetic actuation from manipulator control, thereby simplifying the design of electromagnet's controller. A Soft Actor–Critic (SAC) algorithm~\cite{haarnoja2018soft} is employed to control the electromagnet array.
The controller is first pre-trained in a physics-based simulation environment, and then fine-tuned on the real platform through a sim-to-real training pipeline\cite{zhao2020sim}. An effective control policy can be obtained within approximately 45 minutes (30 minutes model training and 15 minutes setup+sim2real), drastically reducing system setup time.

We validated the platform through trajectory tracking experiments, where a 7-mm magnetic capsule followed 2D paths (square, circular, and long curve). The DRL-based controller outperformed a tuned PID and fix current control baseline, achieving RMSE values of 1.2 mm (square) and 1.5 mm (circular) with real-time performance. The main contributions of this paper are:
\begin{itemize}
\item The design of a portable, low-cost magnetic manipulation platform leveraging a collaborative robot to achieve a large, reconfigurable workspace suitable for full-colon navigation.
\item A DRL control framework that reduces model training and setup time from days to approximately $45$ minutes.
\item Experimental validation demonstrating millimeter-scale tracking accuracy, showcasing the DRL controller's ability to learn and execute precise maneuvers (see Fig.~\ref{fig:overview}).
\end{itemize}

\section{System Overview}
Built entirely from off-the-shelf electronics and a single 3D-printed frame,
the system avoids custom PCBs and room-scale motion-capture, while near-field coil
placement reduces current and power-supply requirements. The prototype can be
reproduced with commodity parts for \(\sim\$200\).

\subsection{Mechanical Design}

\begin{figure}[h]
  \centering 

  \begin{subfigure}[t]{0.48\linewidth} 
    \centering
    \includegraphics[height=0.20\textheight,keepaspectratio,
                     trim=70mm 0mm 0mm 0mm,clip]{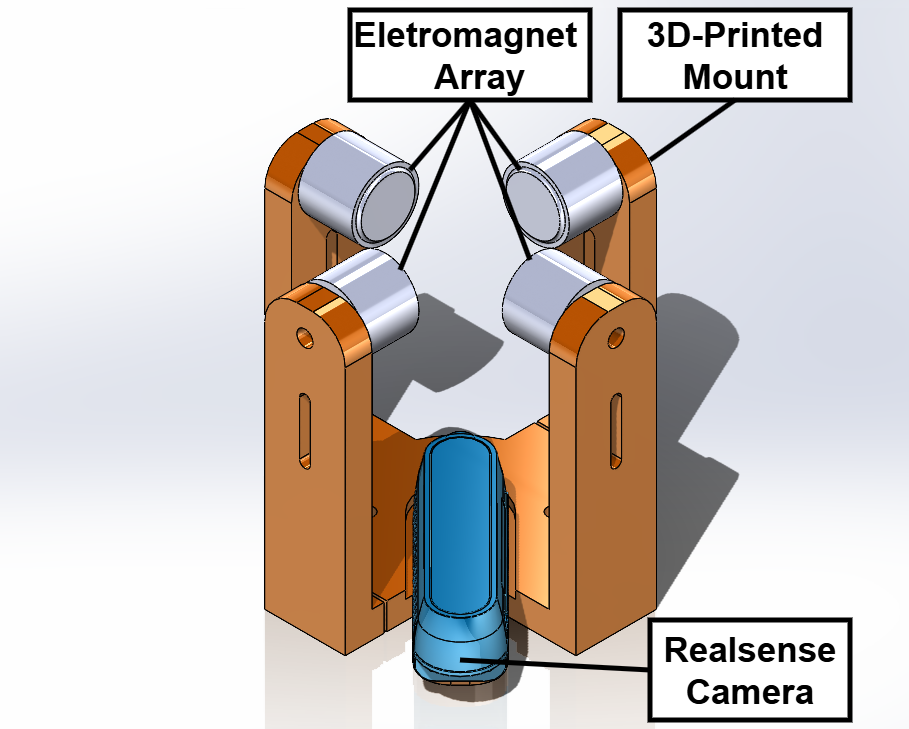}
    \caption{}
    \label{fig:mount}
  \end{subfigure}\hfill
  \begin{subfigure}[t]{0.48\linewidth}
    \centering
    \includegraphics[height=0.20\textheight,keepaspectratio,
                     trim=0mm 0mm 0mm 0mm,clip]{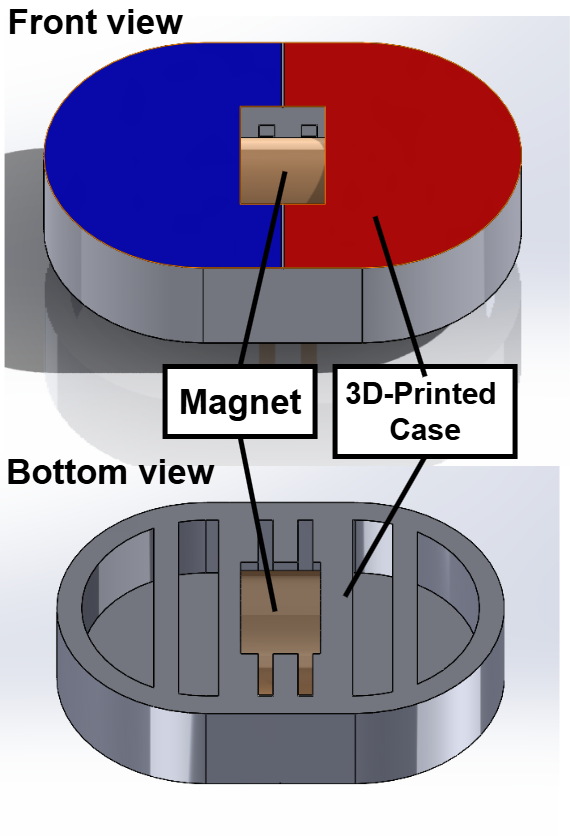}
    \caption{}
    \label{fig:needle}
  \end{subfigure}
  \caption{(a) Electromagnetic end-effector including the four-electromagnet array,
  3D-printed mount and a RealSense camera. (b) Front view and bottom view of the magnetic capsule.}
  \label{fig:mech}
\end{figure}

We enable a large, reconfigurable workspace by mounting a compact four-electromagnet
array on a UR5 collaborative robot. The array, a rigid 3D-printed frame, co-locates the coils with an Intel\texttrademark  RealSense D435 camera. The camera frame is mounted at the UR5 end-effector center, providing a vertical, top-down view of the magnetic agent and its surroundings. The sensing and actuation are then mechanically registered. By moving the array close to the target, we obtain stronger, localized magnetic fields at lower drive while keeping a large, reconfigurable workspace suitable for the GI tract (see Fig.~\ref{fig:mount}). This decoupled design, where robot motion provides coarse positioning and coil currents provide fine translation and rotation, reduces the “model-calibration bottleneck.”  

The magnetic agent in this study is a small permanent-magnetic capsule that floats on the fluid. As shown in Fig.~\ref{fig:needle}, it features a hollow-body design with a bottom cut-out. This structure serves three purposes: the two semicircular bottom cut-outs trap air as buoyant air chambers to keep the capsule at the air–liquid interface; the central bottom hollow functions as a drug reservoir; and the low-friction outer shell limits viscous drag \cite{vella2006load}). This design enables the magnetic capsule to transport therapeutic agents for precise, targeted release at a lesion site, directly fulfilling the drug delivery objective \cite{chen2021triple}.

\subsection{Electromagnetic Actuation System}

\begin{figure}[h] \centering \includegraphics[width=0.7\linewidth, trim=30mm 0mm 5mm 30mm,clip]{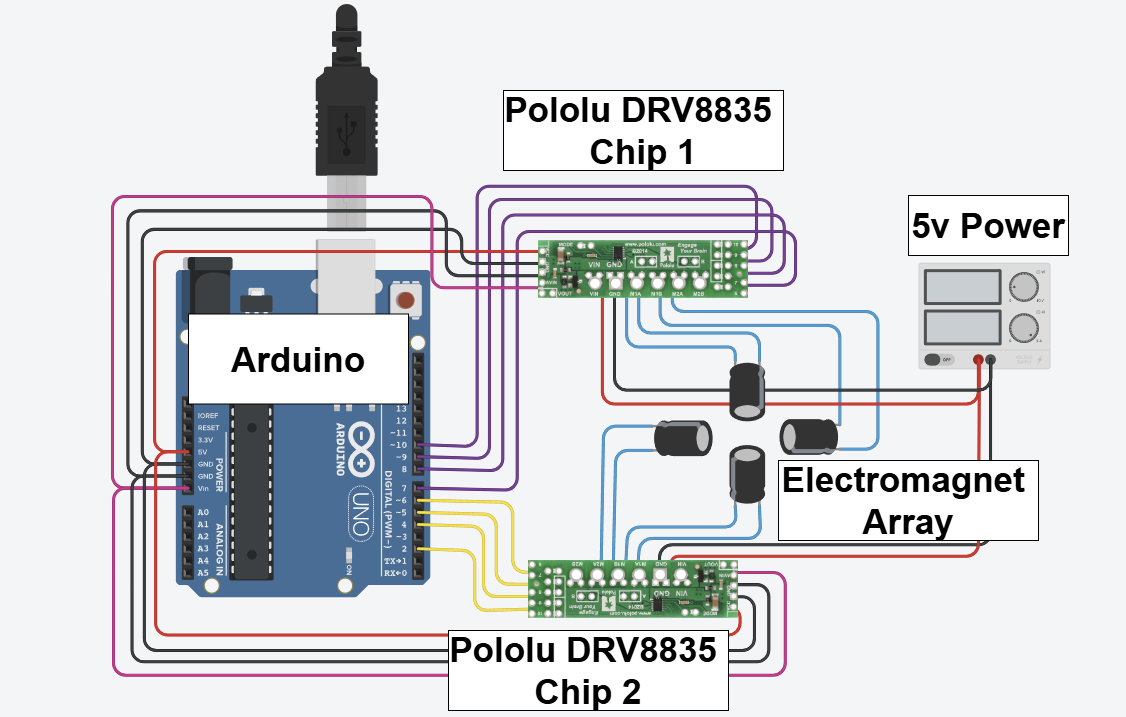} \caption{Schematics of the electromagnet array control. } \label{fig:arduino} \end{figure}

We independently drive a four-coil array built from 5\,V miniature electromagnets
(Adafruit \#3873) using two dual H-bridge boards (DRV8835, Pololu) commanded by
an Arduino Uno R3 (microcontroller). Each driver channel serves one coil; H-bridge
polarity sets current direction and PWM duty cycle sets current magnitude,
enabling fully independent actuation of all four coils  (see Fig.~\ref{fig:arduino}). 

The pose of the magnetic capsule is measured in real-time from the camera's video stream. We apply color thresholding to identify the capsule's distinct red and blue colors, which allows for the precise calculation of its centroid. This pose data is fed to a learning-based controller, which computes the required current magnitudes and polarities to shape the local magnetic field. By adjusting the relative currents across the four coils, the platform generates net forces and torques on the floating magnetic capsule, enabling precise in-plane translation and rotation along the intestine path, without relying on a pre-calibrated magnetic-field model.

\section{Reinforcement Learning–Based Magnetic Tracking Control}

This section presents the DRL framework developed for precise magnetic tracking control.
The framework employs physics-based magnetic \cite{furlani2001permanent} and Fossen models \cite{fossen2011handbook} to construct a simulation environment for training SAC algorithm.
Through simulation pre-training followed by few-shot fine-tuning on the real system, the trained controller efficiently learns the control policy and maintains robust performance under model uncertainties and external disturbances.

\subsection{Problem Formulation}
The objective of the controller is to regulate the planar position and orientation of a magnetic element using four electromagnetic coils. 
Let the system state be defined as
\begin{equation}
x_t = [x,\, y,\, \theta,\, \dot{x},\, \dot{y},\, \dot{\theta}] ,
\end{equation}
and the control input as the coil current vector
\begin{equation}
u_t = [I_1,\, I_2,\, I_3,\, I_4].
\end{equation}

The control objective is to minimize the tracking error between the current and desired states,
\begin{equation}
e_t = x_t - x_d(t),
\end{equation}
where \(x_d(t)\) denotes the desired pose of the magnetic element.

To enable learning-based control, this tracking control problem is reformulated as a Markov Decision Process (MDP)
\begin{equation}
\mathcal{M} = (\mathcal{S},\, \mathcal{A},\, P,\, r,\, \gamma),
\end{equation}
where the state space \(\mathcal{S}\) corresponds to the physical state \(x_t\), 
the action space \(\mathcal{A}\) corresponds to the coil input \(u_t\),  P denotes the state transition distribution,
and the reward function \(r(s_t, a_t)\) is designed based on the tracking performance. 
The discount factor \(\gamma \in (0,1)\) balances immediate and long-term rewards. 
State transitions are sampled from the physics-based simulator or the real environment. 

\subsection{Control Framework}
The proposed control framework leverages physics-based modeling to construct a realistic simulation environment for reinforcement learning. 
The magnetic dipole model \cite{furlani2001permanent} and Fossen model \cite{fossen2011handbook} are integrated to simulate the magnetic field distribution and vehicle–fluid dynamics of the magnetic element.
Within this physic-based simulator, the SAC algorithm is trained to learn a control policy that captures the system’s essential dynamics, resulting in a pre-trained policy.

To bridge the sim-to-real gap, the pre-trained policy is fine-tuned on the mobile magnetic manipulation platform using a few-shot adaptation process with few real-world samples. 
The fine-tuned policy is then deployed on the same platform for real-time magnetic tracking control.

\subsubsection{Physics Models for Simulator}

The hydrodynamic behavior of the capsule is described by the Fossen model, while the magnetic interaction is captured using a four-coil dipole model.

Let $q=[x\ y\ \theta]^{\top}$ and $\dot{q}=[\dot{x}\ \dot{y}\ \dot{\theta}]^{\top}$ denote the planar position, orientation, and their time derivatives, respectively. 
The body-frame velocity is $\nu=[v_x,\ v_y,\ \omega]^{\top}$, related to the inertial-frame velocity through
\begin{equation}
\nu = T(\theta)\dot{q},\quad 
T(\theta)=
\begin{bmatrix}
R^{\top}(\theta) & 0\\
0 & 1
\end{bmatrix},
\end{equation}
where $R(\theta)$ is the body-to-inertial rotation matrix.
The Fossen model in the body frame is
\begin{equation}
M_a\dot{\nu} + C_a(\nu)\nu + D(\nu)\nu + g(q)=\tau_{\text{ext}},
\end{equation}
where $M_a$ is a diagonal inertia matrix, $C_a(\nu)$ represents the Coriolis and centripetal terms, 
and $D(\nu)$ is a diagonal hydrodynamic damping matrix.
For neutrally buoyant planar motion, the restoring term $g(q)$ is negligible.


To compute the magnetic force and torque acting on the capsule, we model the electromagnetic actuation produced by the four coils using the dipole formulation. Each coil $i\in\{1,\dots,4\}$ is modeled as a magnetic dipole with
\begin{equation}
\mathbf{m}_i=\kappa_i I_i \hat{\mathbf{n}}_i,
\end{equation}
where $I_i$ is the coil current, $\hat{\mathbf{n}}_i$ is the coil axis, and $\kappa_i$ is a calibration coefficient. 
Let the capsule center be $\mathbf{r}=[x\ y]^{\top}$, and define $\mathbf{r}_i=\mathbf{r}-\mathbf{c}_i$, $r_i=\|\mathbf{r}_i\|$, and $\hat{\mathbf{r}}_i=\mathbf{r}_i/r_i$, where $c_i$ denotes the position vector of the $i$-th coil. 
The magnetic field generated by coil $i$ is
\begin{equation}
\mathbf{B}_i(\mathbf{r})=
\frac{\mu_0}{4\pi r_i^3}
\!\left(3(\mathbf{m}_i\!\cdot\!\hat{\mathbf{r}}_i)\hat{\mathbf{r}}_i-\mathbf{m}_i\right),
\quad 
\mathbf{B}(\mathbf{r})=\sum_{i=1}^4\mathbf{B}_i(\mathbf{r}).
\end{equation}

The resulting magnetic force and torque on the capsule are
\begin{align}
\mathbf{F} &= \sum_{i=1}^4[\nabla\mathbf{B}_i(\mathbf{r})]^{\top}\mathbf{m}_{PM},\\
\tau_z &= (\mathbf{m}_{PM}\times\mathbf{B}(\mathbf{r}))_z,
\end{align}
and the body-frame wrench applied to the Fossen model is
\begin{equation}
\tau_{\text{ext}}=
\begin{bmatrix}
F_x^b\\ F_y^b\\ \tau_z
\end{bmatrix}
=
\begin{bmatrix}
R^{\top}(\theta)\mathbf{F}\\ \tau_z
\end{bmatrix},
\end{equation}
 Together, the Fossen model and the magnetic dipole model form the physics-based simulator used for control and learning.

\subsubsection{Soft Actor-Critic (SAC)}

SAC~\cite{haarnoja2018soft} is a maximum-entropy RL algorithm that balances reward maximization with exploration by optimizing:
\begin{equation}
\pi^{*} = 
\arg\max_{\pi} \mathbb{E}_{T \sim \pi} 
\left[
\sum_{t} r(s_t, a_t) + 
\alpha \mathcal{H}(\pi(\cdot | s_t))
\right],
\end{equation}
where $r(s_t, a_t)$ is the reward, 
$\mathcal{H}(\pi(\cdot | s_t))$ is the entropy of the policy, 
and $\alpha > 0$ is a parameter 
balancing reward maximization and exploration.



Algorithm~\ref{alg:sac_concise} summarizes the SAC algorithm.

\begin{algorithm}[t]
\caption{SAC}
\label{alg:sac_concise}
\small
\begin{algorithmic}[1]
\STATE \textbf{Init} critics $Q_{\omega_1},Q_{\omega_2}$, actor $\pi_\epsilon$, targets $\bar\omega_j\!\leftarrow\!\omega_j$ $(j\!=\!1,2)$, coefficient $\alpha$, replay buffer $\mathcal{D}$
\FOR{episode $e=1\ldots E$}
  \STATE reset env, get $s$
  \FOR{timestep $t=1\ldots T$}
    \STATE sample $a\sim\pi_\epsilon(\cdot\!\mid\! s)$,\; step env $\to (r,s')$,\; store $(s,a,r,s')$ in $\mathcal{D}$,\; $s\!\leftarrow\! s'$
    \FOR{update $k=1\ldots K$}
      \STATE sample minibatch $\{(s_i,a_i,r_i,s'_i)\}_{i=1}^N\!\sim\!\mathcal{D}$
      \STATE sample $a'_i\!\sim\!\pi_\epsilon(\cdot\!\mid\! s'_i)$,\; compute target
      \vspace{-0.3em}
      \[
      y_i=r_i+\gamma\!\left[\min_{j=1,2} Q_{\bar\omega_j}(s'_i,a'_i)-\alpha\log\pi_\epsilon(a'_i\!\mid\! s'_i)\right]
      \]
      \STATE \textbf{Critics:}\; $\displaystyle
      \min_{\omega_j}\frac{1}{N}\sum_{i}(Q_{\omega_j}(s_i,a_i)-y_i)^2,\ \ j=1,2$
      \STATE \textbf{Actor:}\; sample $a_i\!\sim\!\pi_\epsilon(\cdot\!\mid\! s_i)$, then
      \[
      \min_{\epsilon}\frac{1}{N}\sum_{i}\!\left[\alpha\log\pi_\epsilon(a_i\!\mid\! s_i)-\min_{j}Q_{\omega_j}(s_i,a_i)\right]
      \]
      \STATE Adjust coefficient $\alpha$
      \STATE $\bar\omega_j\!\leftarrow\!\rho\,\omega_j+(1-\rho)\bar\omega_j,\ \ j=1,2$
    \ENDFOR
  \ENDFOR
\ENDFOR
\end{algorithmic}
\end{algorithm}

\begin{figure}[h]
  \centering
  \includegraphics[width=0.9\linewidth,trim=5mm 5mm 5mm 5mm,clip]{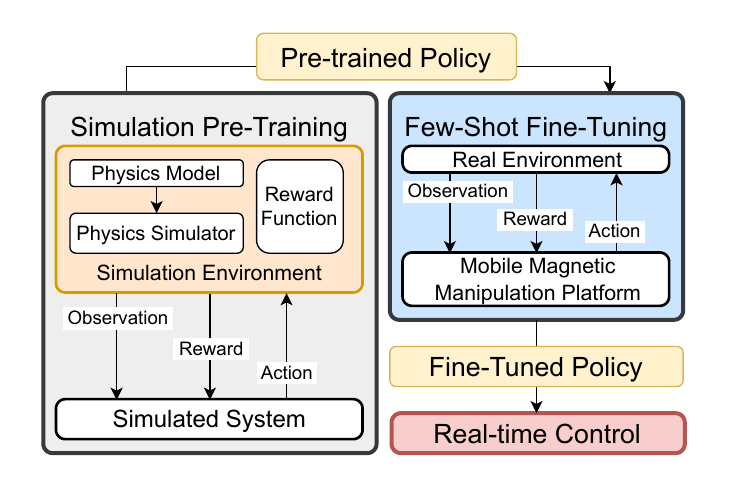}
  \caption{Overview of our sim-to-real pipeline. A policy is pre-trained in a physics-based simulation and then few-shot fine-tuned on our mobile magnetic manipulation platform, yielding a fine-tuned policy for real-time control on the physical system.}
  \label{fig:rl_pipeline}
\end{figure}

\subsubsection{Sim-to-Real Transfer}
To bridge the sim-to-real gap, a two-stage training strategy is applied.
In Stage~1, the SAC algorithm is pre-trained in the physics-based simulator, with domain randomization of coil parameters and damping coefficients to improve generalization.
In Stage~2, few-shot fine-tuning is conducted using limited experimental data to adapt the pre-trained policy to real magnetic and sensing conditions.
This process ensures rapid and reliable policy transfer to the real system (see in Fig.~\ref{fig:rl_pipeline}).

\subsection{Implementation}

The proposed controller was implemented using the Stable-Baselines3 library with a PyTorch backend on Ubuntu 22.04.
All experiments were conducted on a workstation equipped with an NVIDIA RTX~4060.

Training was divided into two phases.
In the simulation environment, model-based SAC training lasted approximately 30 minutes,
while in the real environment, few-shot fine-tuning required about 15 minutes.
GPU acceleration significantly reduced convergence time, and both phases were executed at a fixed control frequency to ensure consistent policy updates.


The composite reward was designed to encourage the magnetic element to approach the target and remain stable after convergence.
It included a distance penalty to reduce position error, a directional reward for velocity alignment, and a progress term for continuous approach.
Near the goal, proximity and stability bonuses reinforced precise convergence, while laziness, smoothness, and energy penalties discouraged weak or inefficient control.
A terminal bonus was given upon successful stabilization.

The main hyperparameters were configured as follows: 
the learning rate was set to $1\times10^{-4}$ in simulation and $5\times10^{-5}$ in real-world fine-tuning; 
the replay buffer size was 185{,}000, with 2{,}000 warm-up steps, 
a batch size of 360, and a discount factor of $\gamma = 0.92$.


\section{Experiments And Results}

We conduct a series of experiments to validate the performance of our learning-based control framework. The objectives are threefold: (1) To quantify the tracking accuracy of our framework, which operates without an explicit, pre-calibrated magnetic field model; (2) To benchmark the DRL controller against traditional (eg. PID) and baseline control strategies (fix current control); and (3) To demonstrate the system's capability for precise, large-scale trajectory following within a clinically-relevant workspace.

\subsection{Experimental Setup}

All experiments were performed using the mobile magnetic manipulation platform described in Section I. Unless otherwise noted, experiments used a water–glycerol medium maintained at \SI{37}{\celsius} and adjusted to a gastrointestinal-relevant viscosity of 2--3~mPa·s  \cite{pedersen2013characterization}, with viscosity computed from the glycerol–water correlation in \cite{cheng2008formula}. This formulation approximates gastric fluid properties while minimizing nonhydrodynamic confounders. We then implemented and evaluated three distinct control strategies:

\begin{itemize}
    \item \textbf{Fixed Current Control (FCC)}: This baseline simulated a static magnetic field, analogous to control with a fixed permanent magnet. We applied a constant current vector, determined by the average current output by the trained DRL control policy when holding the magnetic capsule at a stable equilibrium. 
    
    \item \textbf{PID Control}: A conventional feedback controller based on the analytical dipole model (Section II-B-1). The controller gains (Proportional, Integral, Derivative) were manually tuned for the best possible performance. 
    
    \item \textbf{DRL Control (Ours)}: The proposed control framework described in Section II. 
\end{itemize}

To quantitatively assess tracking performance, we used the Euclidean distance between the magnetic capsule's centroid and the closest point on the reference trajectory as the primary position error metric. Additionally, we evaluated orientation accuracy by tracking a vector aligned with the capsule's longitudinal axis, directed from the red to the blue section. For both metrics, we computed the Root-Mean-Square Error (RMSE), Standard Deviation (Std. Dev.), and Maximum (Max) error relative to the desired trajectory. All comparative results are averaged over five independent trials.

\subsection{Comparative Analysis of Trajectory Tracking}

In the first set of experiments, we tasked the three controllers with tracking two standard 2D trajectories: a 40$\times$40 mm square and a circle with a 20 mm radius. These paths were chosen to evaluate the controllers' ability to handle both sharp turns (square) and smooth curves (circle). Additionally, the desired orientation was defined to be tangential to the trajectory for the square path, and perpendicular to the trajectory for the circular path.

\begin{figure}[t]
  \centering

  \newcommand{\leftw}{0.4\columnwidth}
  \newcommand{\rightw}{0.49\columnwidth} 
  \newcommand{\colgap}{0\columnwidth}

  \begin{minipage}[t]{\leftw}
    \centering
    \includegraphics[width=\linewidth,clip,trim=0mm 0mm 0mm 0mm]{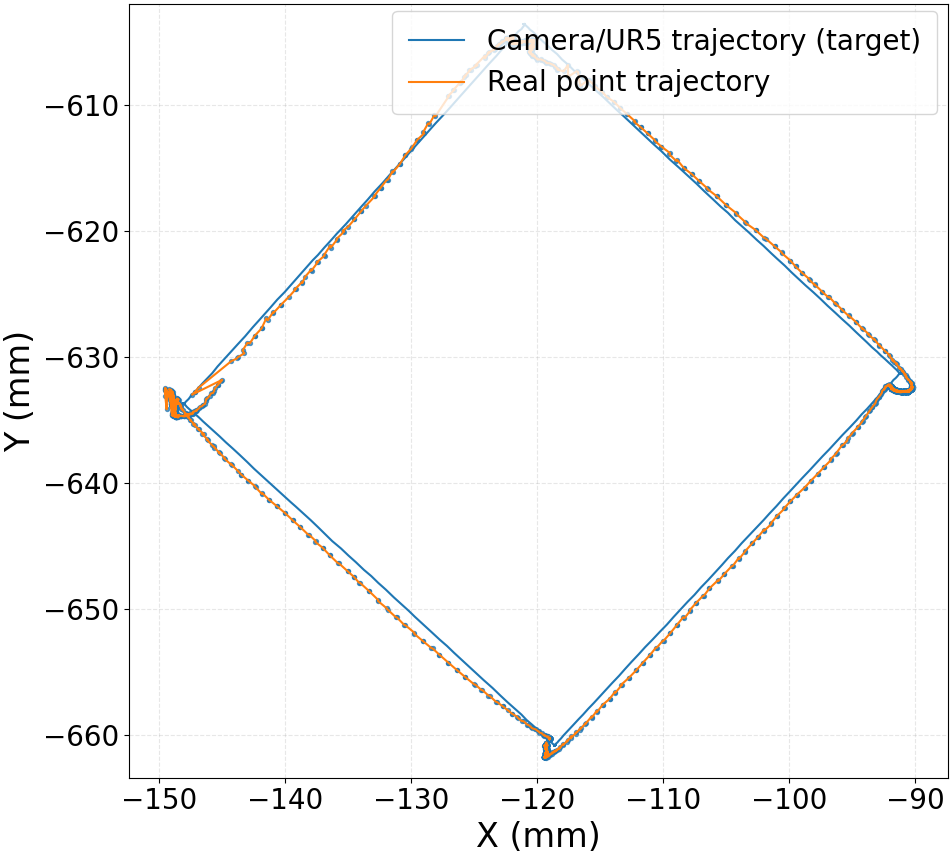}
    \caption*{(a)}
  \end{minipage}\hspace{\colgap}%
  \begin{minipage}[t]{\rightw}
    \centering
    \includegraphics[width=\linewidth,clip,trim=0mm 4mm 0mm 0mm]{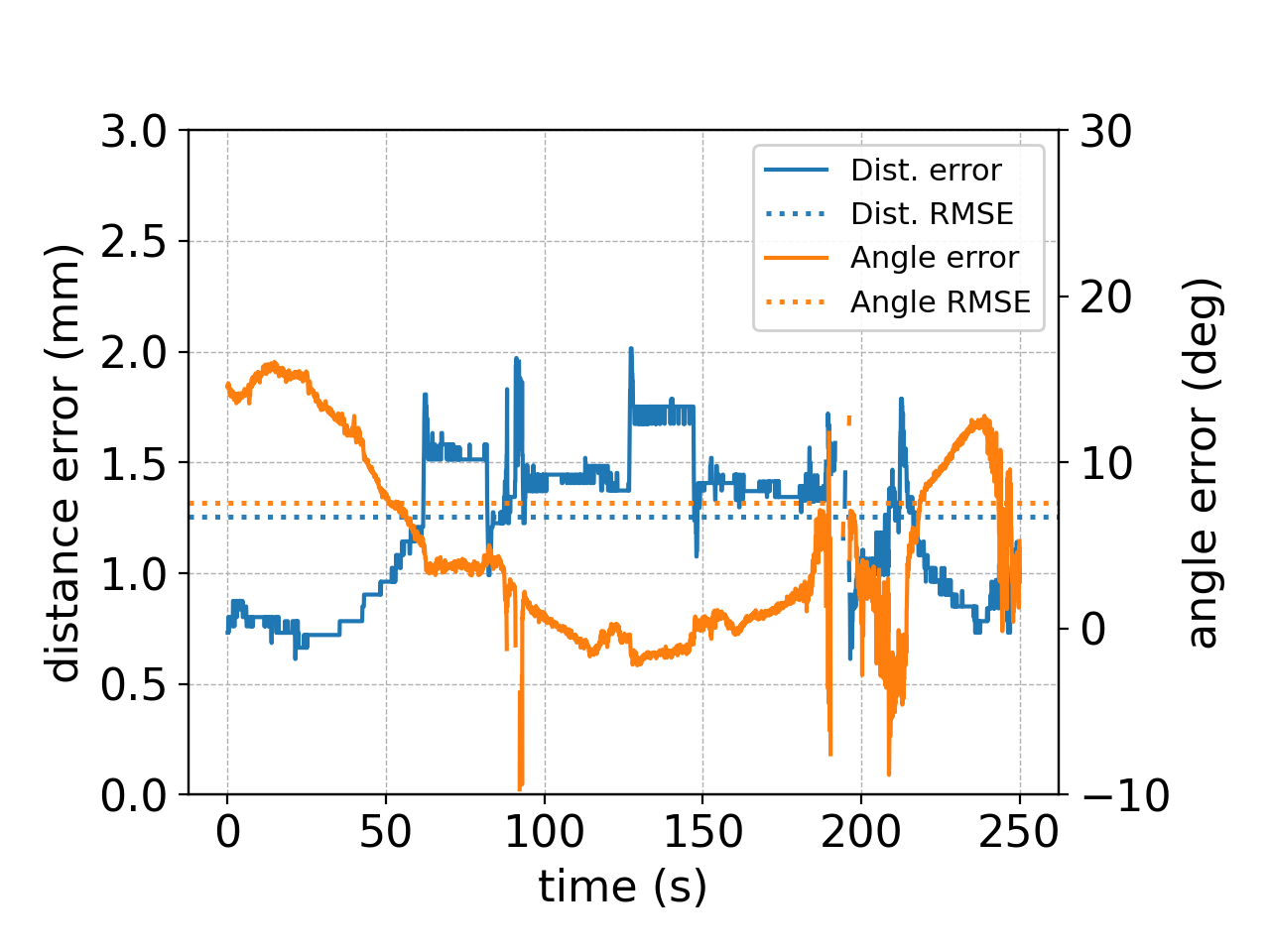}
    \caption*{(b)}
  \end{minipage}

  \vspace{6pt}

  \begin{minipage}[t]{\leftw}
    \centering
    \includegraphics[width=\linewidth,clip,trim=0mm 0mm 0mm 0mm]{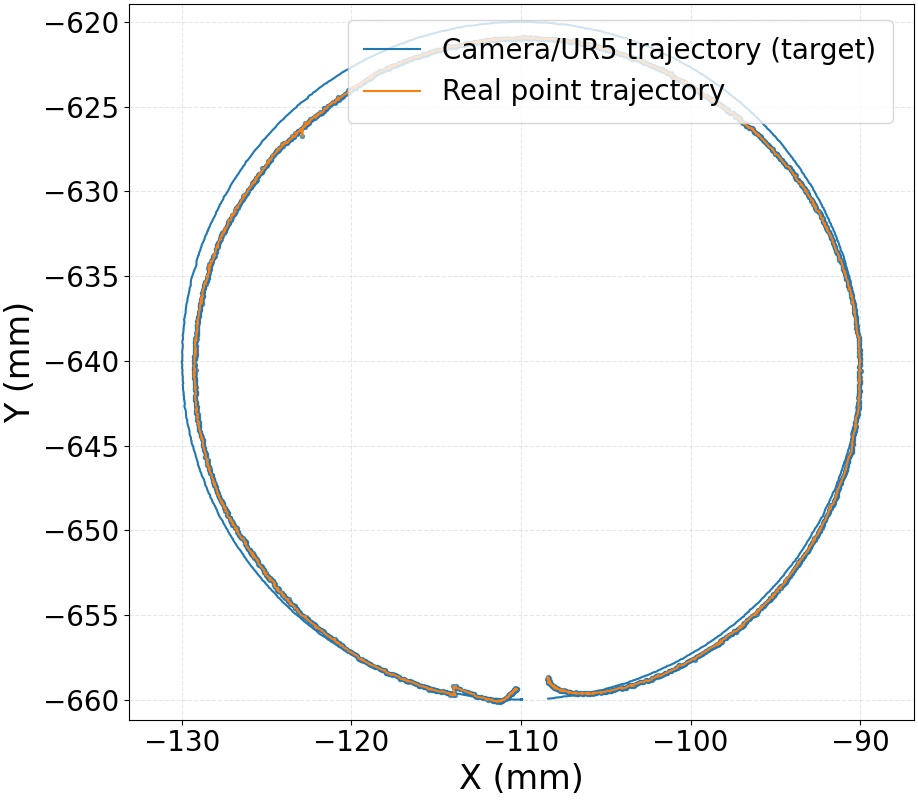}
    \caption*{(c)}
  \end{minipage}\hspace{\colgap}%
  \begin{minipage}[t]{\rightw}
    \centering
    \includegraphics[width=\linewidth,clip,trim=0mm 4mm 0mm 0mm]{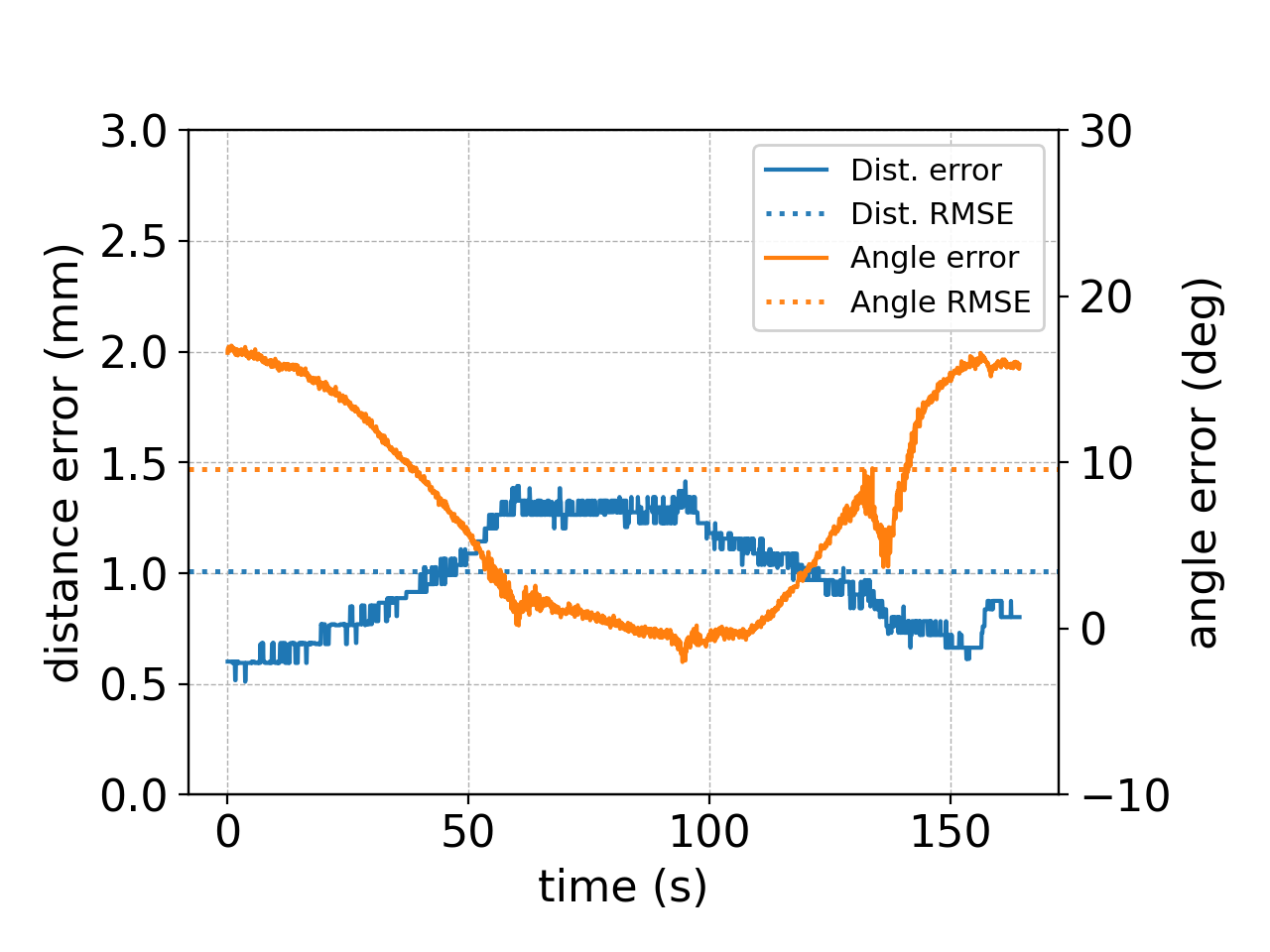}
    \caption*{(d)}
  \end{minipage}

  \caption{DRL Controller tracking performance for square and circular trajectories showing a representative sample.
  (a) Trajectory tracking of the 40$\times$40\,mm square.
  (b) Error of the square: distance RMSE = 1.01\,mm, angle RMSE = 9.57$^{\circ}$.
  (c) Trajectory tracking of the 20\,mm radius circle.
  (d) Error of the circle: distance RMSE = 1.25\,mm, angle RMSE = 7.59$^{\circ}$.}
  \label{fig:path_comparison}
  \vspace{-2mm}
\end{figure}

The quantitative results are summarized in Table~\ref{tab:transposed_results}, and the tracked paths are visualized in Fig.~\ref{fig:path_comparison}. The analysis of these results is as follows:
\begin{table}[h]
\caption{Consolidated tracking performance comparison (average of 5 Trials). }
\label{tab:transposed_results} 
\centering
\small 
\setlength{\tabcolsep}{2pt} 

\sisetup{
  table-format=-2.2, 
  detect-weight,
  mode=text
}

\begin{tabular}{
  l l 
  S 
  S 
  S 
}
\toprule 
\textbf{Path} & \textbf{Metric} & {\textbf{FCC}} & {\textbf{PID}} & {\textbf{DRL}} \\
\midrule

\multirow{6}{*}{Square} 
  & Dist. RMSE (mm) & 2.64 & 4.72 & \bfseries 1.18 \\
  & Dist. Std. (mm) & 1.81 & 1.28 & \bfseries 0.34 \\
  & Dist. Max (mm)  & 16.00 & 8.46 & \bfseries 2.05 \\
 \cmidrule(lr){2-5} 
  & Angle RMSE ($^{\circ}$) & 15.10 & {---} & \bfseries 9.24 \\
  & Angle Std. ($^{\circ}$) & 12.79 & {---} & \bfseries 6.56 \\
  & Angle Max ($^{\circ}$)  & 96.07 & {---} & \bfseries 32.56 \\
\midrule

\multirow{6}{*}{Circle} 
  & Dist. RMSE (mm) & 2.61 & 4.75 & \bfseries 1.50 \\
  & Dist. Std. (mm) & 1.45 & 0.94 & \bfseries 0.31 \\
  & Dist. Max (mm)  & 15.91 & 8.26 & \bfseries 2.47 \\
 \cmidrule(lr){2-5} 
  & Angle RMSE ($^{\circ}$) & 13.81 & {    ---} & \bfseries 10.47 \\
  & Angle Std. ($^{\circ}$) & 9.53 & {    ---} & \bfseries 7.63 \\
  & Angle Max ($^{\circ}$)  & 57.25 & {    ---} & \bfseries 29.81 \\
\bottomrule
\end{tabular}
\end{table}

\begin{itemize}
    \item \textbf{DRL Control}: The proposed control
framework consistently achieved the lowest RMSE and standard deviation, demonstrating superior accuracy and robustness. It effectively developed a control policy capable of stabilizing the capsule at a preferred location while successfully navigating diverse trajectories, including both sharp turns and smooth curves (Table~\ref{tab:transposed_results}).
    
    \item \textbf{Fixed Current Control (FCC)}: This passive baseline produced a quasi-static pulling field. It lacked active compensation and could not reject disturbances or regulate orientation, leading to sluggish response and decoupling at higher speeds or sharp turns.

    \item \textbf{PID Control}: The PID controller underperformed, with RMSE values worse than even the FCC baseline. The main cause was severe model mismatch under an uncalibrated system: without a full-system calibration (field–current map, coil alignment, and the capsule’s magnetic moment/orientation). Consequently, we could only implement a \emph{position-only} PID loop; a reliable orientation controller was infeasible. This limitation produced corner overshoot, oscillations, and steady-state jitter. Angle metrics for PID were therefore not reported (---).
\end{itemize}

Given the clear superiority in both accuracy and stability, the DRL Controller was exclusively selected for the clinically-relevant navigation task.

\subsection{Clinically Relevant Workspace Validation}

\begin{figure}[h]
  \centering
  \includegraphics[
    width=0.7\linewidth,
    origin=c,
    clip,
    trim=12mm 6mm 10mm 8mm 
  ]{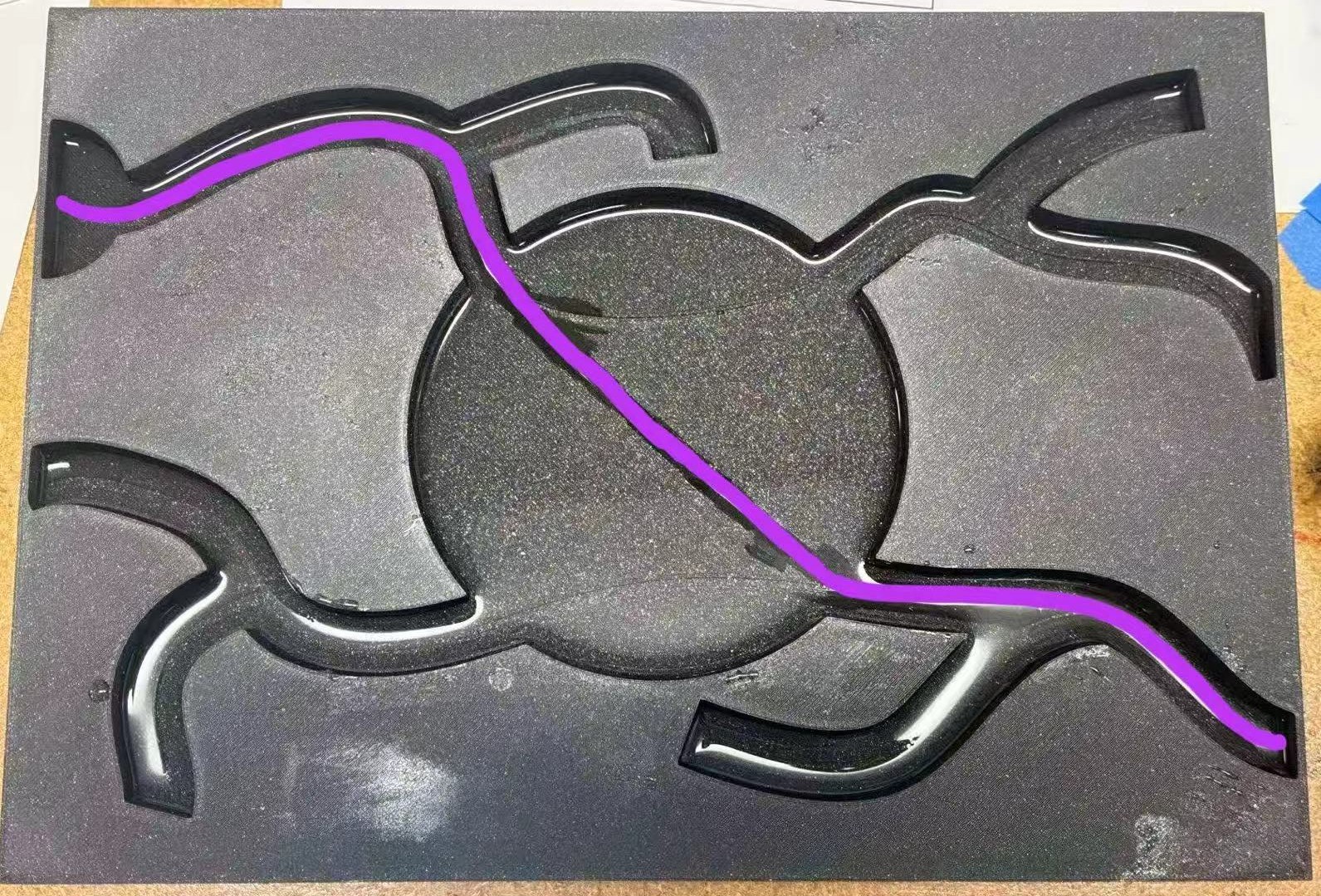}
  \caption{Continuous trajectory spanning
a 30 cm × 20 cm area, the purple line is the desired trajectory. }
  \label{fig:long_path}
\end{figure}

We validated the system's performance in a clinically-relevant scenario using a large 30 cm $\times$ 20 cm trajectory, simulating a significant colon section. The path included complex curves and narrow passages with a 10~mm minimum width. This minimal width was chosen to be more challenging than the average healthy colonic lumen (2.5–7 cm \cite{liao2009gastrointestinal}), thereby simulating navigation through a difficult, partially obstructed, or pathological region.(see Fig~\ref{fig:long_path}) This task requires the UR5 robotic arm to move and reposition the end-effector to keep the 7~mm magnetic capsule within its effective actuation range, while the DRL Controller continuously manages coil currents for fine-grained control.

Across three repeated trials, the DRL Controller successfully maintained high-fidelity tracking throughout this complex workspace. As shown in Table~\ref{tab:traj_detail}, the system achieved an average tracking RMSE of 1.50~mm. This experiment confirms that our DRL-based approach, combined with a mobile manipulator, successfully overcomes the workspace limitations of fixed-coil systems without succumbing to the model-calibration bottleneck, demonstrating precise, millimeter-scale control in a large and challenging phantom.

\begin{table}[h]
\caption{Detailed trial data for DRL continuous trajectory.}
\label{tab:traj_detail}
\centering
\footnotesize 
\setlength{\tabcolsep}{3pt}

\sisetup{
  table-format=2.2, 
  detect-weight,
  mode=text
}

\begin{tabular}{
    l 
    S[table-format=1.2] 
    S[table-format=1.2] 
    S[table-format=1.2] 
    S[table-format=2.2] 
    S[table-format=1.2] 
    S[table-format=2.2] 
}
\toprule 

& \multicolumn{3}{c}{\textbf{Distance Error (mm)}} & \multicolumn{3}{c}{\textbf{Angle Error ($^{\circ}$)}} \\
\cmidrule(lr){2-4} \cmidrule(lr){5-7}
\textbf{Trial} & {\textbf{RMSE}} & {\textbf{Std.}} & {\textbf{Max}} & {\textbf{RMSE}} & {\textbf{Std.}} & {\textbf{Max}} \\
\midrule

Trial 1 & 1.56 & 0.32 & 2.26 & 10.14 & 7.56 & 20.69 \\
Trial 2 & 1.43 & 0.33 & 2.73 & 10.44 & 8.10 & 39.30 \\
Trial 3 & 1.52 & 0.29 & 2.18 & 10.83 & 7.17 & 19.97 \\

\midrule 

\bfseries Average & \bfseries 1.50 & \bfseries 0.31 & \bfseries 2.47 & \bfseries 10.47 & \bfseries 7.63 & \bfseries 29.81 \\
\bottomrule
\end{tabular}
\end{table}


\section{Conclusions}

We presented a portable, low-cost magnetic manipulation platform using a UR5 and a DRL control policy, eliminating the model-calibration bottleneck. Our system achieves millimeter-scale tracking ($\approx$ 1.0–1.5 mm RMSE) over a large, reconfigurable workspace, outperforming PID and fixed-current methods. The entire system, including DRL training, can be deployed in approximately 45 minutes.

The primary limitations of this work are the validation in a 2D rigid phantom and the reliance on camera-based guidance, which is not clinically viable for GI navigation. Future work must integrate clinically-relevant imaging, such as ultrasound, to handle occlusions. We will also focus on migrating to 3D navigation and validating the system in more realistic ex vivo models for targeted gastrointestinal interventions.





\section*{ACKNOWLEDGMENT}

This research has been generously funded by the National Science Foundation’s Foundational Research in Robotics CAREER
program under award number 2144348. The views and opinions expressed in this work are solely
those of the authors and do not necessarily reflect the official stance of the 
National Science Foundation.

\clearpage
\bibliographystyle{IEEEtran}
\bibliography{reference}

\end{document}